\documentclass[11pt,a4paper]{article}

\pdfoutput=1

\usepackage{authblk}

\usepackage[hyperref]{emnlp-ijcnlp-2019}
\usepackage{times}
\usepackage{latexsym}
\usepackage{multirow}
\usepackage{bbding}
\usepackage{xcolor}

\aclfinalcopy

\title{Can a Gorilla Ride a Camel? \\ Learning Semantic Plausibility from Text}

\author[1]{Ian Porada}
\author[2]{Kaheer Suleman}
\author[1,3]{Jackie Chi Kit Cheung}
\affil[1]{Mila, McGill University}
\affil[ ]{{\tt \{ian.porada@mail, jcheung@cs\}.mcgill.ca}}
\affil[2]{Microsoft Research Montreal}
\affil[ ]{{\tt kasulema@microsoft.com}}
\affil[3]{Canada CIFAR AI Chair}
\date{}

\begin{document}
\maketitle
\begin{abstract}
  Modeling semantic plausibility requires commonsense knowledge about
  the world and has been used as a testbed for exploring various knowledge representations. 
  Previous work has focused specifically
  on modeling physical plausibility and shown that distributional
  methods fail when tested in a supervised setting. At the same time,
  distributional models, namely large pretrained language models, have led to improved results for
  many natural language understanding tasks. In this work, we show
  that these pretrained language models are in fact effective at
  modeling physical plausibility in the supervised setting. We
  therefore present the more difficult problem of learning to model
  physical plausibility directly from text. We create a training set
  by extracting attested events from a large corpus, and we provide a
  baseline for training on these attested events in a self-supervised
  manner and testing on a physical plausibility task. We believe results
  could be further improved by injecting explicit commonsense knowledge
  into a distributional model.
\end{abstract}

\section{Introduction}

A person riding a camel is a common event, and one would
expect the subject-verb-object (s-v-o) triple
\textit{person-ride-camel} to be attested in a large corpus. In
contrast, \textit{gorilla-ride-camel} is uncommon, likely 
unattested, and yet still semantically plausible. Modeling semantic
plausibility then requires distinguishing these plausible events from
the semantically nonsensical, e.g. \textit{lake-ride-camel}.

Semantic plausibility is a necessary part of many natural
language understanding (NLU) tasks including narrative interpolation \cite{bowman-etal-2016-generating}, story understanding \cite{mostafazadeh-etal-2016-corpus}, paragraph reconstruction \cite{li-jurafsky-2017-neural}, and hard coreference resolution \cite{peng-etal-2015-solving}.
Furthermore, the problem of modeling semantic plausibility has itself been used as a testbed for exploring various knowledge representations.

\begin{table}[t]
\begin{center}
\def\arraystretch{1.25}
\begin{tabular}{c|c}
{\bf Event} & {\bf Plausible?} \\
\hline {\it bird-construct-nest} &  \Checkmark \\
{\it bottle-contain-elephant} & \XSolidBrush \\
{\it gorilla-ride-camel} &  \Checkmark \\
{\it lake-fuse-tie} &  \XSolidBrush \\
\end{tabular}
\end{center}
\caption{\label{samples} Example events from \citet{wang-etal-2018-modeling}'s physical plausibility dataset. }
\end{table}

In this work, we focus specifically on modeling physical plausibility as presented by \citet{wang-etal-2018-modeling}. This is the problem of determining if a given event, represented as an s-v-o triple, is physically plausible (Table~\ref{samples}). We show that in the original supervised setting a distributional model, namely a novel application of BERT \cite{devlin-etal-2019-bert}, significantly outperforms the best existing
method which has access to manually labeled physical features \cite{wang-etal-2018-modeling}.

Still, the generalization ability of supervised models is limited by
the coverage of the training set. We therefore present the more difficult problem of learning physical plausibility
directly from text. We create a training set by parsing and
extracting attested s-v-o triples from English Wikipedia, and we provide a
baseline for training on this dataset and evaluating on
\citet{wang-etal-2018-modeling}'s physical plausibility task. We also
experiment training on a large set of s-v-o triples extracted from
the web as part of the NELL project
\cite{Carlson:2010:TAN:2898607.2898816}, and find that Wikipedia
triples result in better performance.

\section{Related Work}

\citet{wang-etal-2018-modeling} present the semantic plausibility
dataset that we use for evaluation in this work, and they show that
distributional methods fail on this dataset. This conclusion aligns with
other work showing that GloVe \cite{pennington-etal-2014-glove} and
word2vec \cite{mikolov2013distributed} embeddings do not encode some
salient features of objects \cite{li-gauthier-2017-distributional}.
More recent work has similarly concluded that large pretrained language models
only learn attested physical knowledge \cite{forbes2019neural}.

Other datasets which include plausibility ratings are smaller
in size and missing atypical but plausible events \cite{Keller:2003},
or concern the more complicated problem of multi-event inference in
natural language \cite{zhang-etal-2017-ordinal,sap2019atomic}.

Complementary to our work are methods of extracting physical
features from a text corpus
\cite{wang-etal-2017-distributional,forbes-choi-2017-verb,Bagherinezhad2016AreEB}.

\subsection{Distributional Models}

Motivated by the distributional hypothesis that
words in similar contexts have similar meanings
\cite{harris_doi:10.1080/00437956.1954.11659520}, distributional methods learn the
representation of a word based on the distribution of its context.
The occurrence counts of bigrams in a corpus are correlated with
human plausibility ratings
\cite{Lapata:1999:DAP:977035.977041,Lapata:2001:ESA:1073012.1073058},
so one might expect that with a large enough corpus, a distributional
model would learn to distinguish plausible but atypical events from
implausible ones. As a counterexample,
\citet{Seaghdha:2010:LVM:1858681.1858726} has shown that the
subject-verb bigram \textit{carrot-laugh} occurs 855 times in a web
corpus, while \textit{manservant-laugh} occurs zero.\footnote{This
point was made based on search engine results. Some, but not all, of
the \textit{carrot-laugh} bigrams are false positives.} Not
everything that is physically plausible occurs, and not everything
that occurs is attested due to reporting bias\footnote{Reporting bias
describes the discrepancy between what is frequent in text and what
is likely in the world. This is in part because people do not
describe the obvious.} \cite{Gordon:2013:RBK:2509558.2509563}; therefore, modeling semantic plausibility requires systematic inference beyond a distributional cue.

We focus on the masked language model BERT as a distributional model. BERT has led to improved results across a variety of NLU benchmarks \cite{rajpurkar2018know,wang2018glue}, including tasks that require explicit commonsense reasoning such as the Winograd Schema Challenge \cite{Sakaguchi2019WINOGRANDEAA}.

\subsection{Selectional Preference}

Closely related to semantic plausibility is selectional preference
\cite{resnik1996selectional} which concerns the semantic preference
of a predicate for its arguments. Here, \textit{preference} refers to
the typicality of arguments: while it is plausible that a gorilla
rides a camel, it is not preferred. Current approaches to
selectional preference are distributional
\cite{erk-etal-2010-flexible,van-de-cruys-2014-neural} and have shown limited performance in
capturing semantic plausibility \cite{wang-etal-2018-modeling}. 

\citet{o-seaghdha-korhonen-2012-modelling} have investigated combining
a lexical hierarchy with a distributional approach, and there have been related attempts at grounding selectional preference in
visual perception \cite{bergsma-goebel-2011-using,shutova-etal-2015-perceptually}.

Models of selectional preference are either evaluated on a pseudo-disambiguation
task, where attested predicate-argument tuples must be
disambiguated from pseudo-negative random tuples, or evaluated on
their correlation with human plausibility judgments. Selectional
preference is one factor in plausibility and thus the two should
correlate.

\section{Task}

Following existing work, we focus on the task of single-event,
physical plausibility. This is the problem of determining if a given
event, represented as an s-v-o triple, is physically plausible.

We use \citet{wang-etal-2018-modeling}'s physical plausibility dataset for evaluation. This dataset consists of 3,062 s-v-o triples, built from a vocabulary of 150 verbs and 450 nouns, and containing a diverse combination of both typical and atypical events balanced between the plausible and implausible categories. The set of events and ground truth labels were manually curated.

\subsection{Supervised}

In the supervised setting, a model is trained and tested on labelled events from the same distribution. Therefore, both the training and test set capture typical and atypical plausibility. We follow the same evaluation procedure as previous work and perform cross validation on the 3,062 labeled triples \cite{wang-etal-2018-modeling}.

\begin{table}[t]
\begin{center}
\begin{tabular}{|c|c|}
\hline \multirow{3}{4.5em}{\bf Wikipedia} & {\it male-have-income} \\
& {\it village-have-population} \\
& {\it event-take-place} \\ \hline
\multirow{3}{4.5em}{\bf NELL} & {\it login-post-comment} \\
& {\it use-constitute-acceptance} \\
& {\it modules-have-options} \\ \hline
\end{tabular}
\end{center}
\caption{\label{corpora} Most frequent s-v-o triples for each corpus.}
\end{table}

\subsection{Learning from Text} \label{learning-from-text}

We also present the problem of learning to model physical plausibility directly from text. In this new setting, a model is trained on events extracted from a large corpus and evaluated on a physical plausibility task. Therefore, only the test set covers both typical and atypical plausibility.

We create two training sets based on separate corpora: first, we parse English Wikipedia using the StanfordNLP neural pipeline \cite{qi2018universal} and extract attested s-v-o triples. Wikipedia has led to relatively good results for selectional preference \cite{zhang-etal-2019-sp}, and in total we extract 6 million unique triples with a cumulative 10 million occurrences. Second, we use the NELL \cite{Carlson:2010:TAN:2898607.2898816} dataset of 604 million s-v-o triples extracted from the dependency parsed ClueWeb09 dataset. For NELL, we filter out triples with non-alphabetic characters or less than 5 occurrences, resulting in a total 2.5 million unique triples with a cumulative 112 million occurrences.

For evaluation, we split \citet{wang-etal-2018-modeling}'s 3,062 triples into equal sized validation and test sets. Each set thus consists of 1,531 triples.

\section{Methods}

\subsection{NN}

As a baseline, we consider the performance of a neural method for selectional preference \cite{van-de-cruys-2014-neural}. This method is a two-layer artificial neural network (NN) over static embeddings.

\paragraph{Supervised.} We reproduce the results of \citet{wang-etal-2018-modeling} using GloVe embeddings and the same hyperparameter settings.

\paragraph{Self-Supervised.} We use this same method for \textit{learning from text} (Subsection~\ref{learning-from-text}). To do so, we turn the training data into a self-supervised training set: attested events are considered to be plausible, and pseudo-implausible events are created by sampling each word in an s-v-o triple independently by occurrence frequency. We do hyperparameter search on the validation set over learning rates in $\{1e-3, 1e-4, 1e-5, 2e-5\}$, batch sizes in $\{16, 32, 64, 128\}$, and epochs in $\{0.5, 1, 2\}$. 

\subsection{BERT}

We use BERT for modeling semantic plausibility by simply treating this as a sequence classification task. We tokenize the input s-v-o triple and introduce new entity marker tokens to separate each word.\footnote{Our input to BERT is of the form: \texttt{[CLS] [SUBJ] <subject> [/SUBJ] [VERB] <verb> [/VERB] [OBJ] <object> [/OBJ] [SEP]}.} We then add a single layer NN to classify the input based on the final layer representation of the \texttt{[CLS]} token. We use BERT-large and finetune the entire model in training.\footnote{We use Hugging Face's PyTorch implementation of BERT, \url{https://github.com/huggingface/pytorch-transformers}.}

\paragraph{Supervised.} We do no hyperparameter search and simply use the default hyperparameter configuration which has been shown to work well for other commonsense reasoning tasks \cite{DBLP:journals/corr/abs-1904-09705}. BERT-large sometimes fails to train on small datasets \cite{devlin-etal-2019-bert,niven-kao-2019-probing}; therefore, we restart training with a new random seed when the training loss fails to decrease more than 10\%.

\paragraph{Self-Supervised.} We perform \textit{learning from text} (Subsection~\ref{learning-from-text}) by creating a self-supervised training set in exactly the same way as for the NN method. The hyperparameter configuration is determined by grid search on the validation set over learning rates in $\{1e-5, 2e-5, 3e-5\}$, batch sizes in $\{8, 16\}$, and epochs in $\{0.5, 1, 2\}$.

\section{Results}

\subsection{Supervised}

\begin{table}[t]
\begin{center}
\begin{tabular}{|l|l|}
\hline \bf Model & \bf Accuracy \\ \hline
Random & 0.50 \\
NN \cite{van-de-cruys-2014-neural} & 0.68 \\
NN+WK \cite{wang-etal-2018-modeling} & 0.76 \\
Fine-tuned BERT & \bf 0.89 \\
\hline
\end{tabular}
\end{center}
\caption{\label{supervised} Mean accuracy of classifying plausible events for models trained in a supervised setting. NN+WK combines the NN approach with manually labeled world knowledge (WK) features describing both the subject and object.}
\end{table}

\begin{table}[t]
\begin{center}
\def\arraystretch{1.25}
\begin{tabular}{|c|c|c|}
\hline \multirow{2}{1em}{\bf Event} & \multicolumn{2}{|c|}{\bf Plausible?} \\ \cline{2-3} 
& BERT & GT  \\ \hline
{\it dentist-capsize-canoe} & \Checkmark & \Checkmark \\
{\it stove-heat-air} & \XSolidBrush & \Checkmark \\
{\it sun-cool-water} & \Checkmark & \XSolidBrush \\
{\it chair-crush-water} & \XSolidBrush & \XSolidBrush \\ \hline
\end{tabular}
\end{center}
\caption{\label{confidence} Interpreting log-likelihood as confidence, example events for which BERT was highly confident and either correct or incorrect with respect to the ground truth (GT) label.}
\end{table}

For the supervised setting, we follow the same evaluation procedure as \citet{wang-etal-2018-modeling}: we perform 10-fold cross validation on the dataset of 3,062 s-v-o triples, and report the mean accuracy of running this procedure 20 times all with the same model initialization (Table~\ref{supervised}).

BERT outperforms existing methods by a large margin, including those with access to manually labeled physical features. We conclude from these results that distributional data does provide a strong cue for semantic plausibility in the supervised setting of \citet{wang-etal-2018-modeling}.

Examples of positive and negative results for BERT are presented in Table~\ref{confidence}. There is no immediately obvious pattern in the cases where BERT misclassifies an event. We therefore consider events for which BERT gave a consistent estimate across all 20 runs of cross-validation. Of these, we present the event for which BERT was most confident.

We note that due to the limited vocabulary size of the dataset, the training set always covers the test set vocabulary when performing 10-fold cross validation. That is to say that every word in the test set has been seen in a different triple in the training set. For example, every verb occurs within 20 triples; therefore, on average a verb in the test set has been seen 18 times in the training set.

Supervised performance is dependent on the coverage of the training set vocabulary \cite{moosavi-strube-2017-lexical}, and it is prohibitively expensive to have a high coverage of plausibility labels across all English verbs and nouns. Furthermore, supervised models are susceptible to annotation artifacts \cite{gururangan-etal-2018-annotation,poliak-etal-2018-hypothesis} and do not necessarily even learn the desired relation, or in fact any relation, between words \cite{Levy2015DoSD}.

This is our motivation for reframing semantic plausibility as a task to be learned directly from text, a new setting in which the training set vocabulary is independent of the test set.

\subsection{Learning from Text}

\begin{table}[t]
\begin{center}
\begin{tabular}{|l|l|l|l|l|}
\hline \multirow{2}{1em}{\bf Model} & \multicolumn{2}{|c|}{\bf Wikipedia} & \multicolumn{2}{|c|}{\bf NELL} \\ \cline{2-5}
& Valid & Test & Valid & Test \\ \hline
Random & 0.50 & 0.50 & 0.50 & 0.50 \\
NN & 0.53 & 0.52 & 0.50 & 0.51 \\
BERT & \bf 0.65 & \bf 0.63 & \bf 0.57 & \bf 0.56 \\
\hline
\end{tabular}
\end{center}
\caption{\label{unsupervised} Accuracy of classifying plausible events for models trained on a corpus in a self-supervised manner.}
\end{table}

For \textit{learning from text} (Subsection~\ref{learning-from-text}), we report both the validation and test accuracies of classifying physically plausible events (Table~\ref{unsupervised}).

BERT fine-tuned on Wikipedia performs the best, although only partially captures semantic plausibility with a test set accuracy of 63\%. Performance may benefit from injecting explicit commonsense knowledge into the model, an approach which has previously been used in the supervised setting \cite{wang-etal-2018-modeling}.

Interestingly, BERT is biased towards labelling events as plausible. For the best performing model, for example, 78\% of errors are false positives.

Models trained on Wikipedia events consistently outperform those trained on NELL which is consistent with our subjective assessment of the cleanliness of these datasets. The baseline NN method in particular seems to learn very little from training on the NELL dataset.

\section{Conclusion}

We show that large, pretrained language models are effective at modeling semantic plausibility in the supervised setting. Supervised models are limited by the coverage of the training set, however; thus, we reframe modeling semantic plausibility as a self-supervised task and present a baseline based on a novel application of BERT.

We believe that self-supervised results could be further improved by incorporating explicit commonsense knowledge, as well as further incidental signals \cite{Roth2017IncidentalSM} from text.

\section*{Acknowledgments}
We would like to thank Adam Trischler, Ali Emami, and Abhilasha Ravichander for useful discussions and comments.
This work is supported by funding from Microsoft Research and resources from Compute Canada. The last author is supported by the Canada CIFAR AI Chair program.

\bibliography{emnlp-ijcnlp-2019}
\bibliographystyle{acl_natbib}

\end{document}